\documentclass[letterpaper]{article}
\usepackage[preprint]{aaai2027}
\usepackage[hyphens]{url}
\usepackage{graphicx}
\usepackage{natbib}
\usepackage{caption}
\usepackage{algorithm}
\usepackage{algorithmic}
\usepackage{booktabs}
\usepackage{amsmath,amssymb}
\newcommand{\OH}{\textsc{OHIRL}}

\title{Online Reward-Punishment Learning from Fixed-Channel Perceptual Event Streams without Environment Rewards}
\author{Zirong Li}
\affiliations{Tiangong University\\\texttt{xingyeyouyv@gmail.com}}
\begin{document}
\maketitle

\begin{abstract}
We study online reward-punishment learning when the environment emits no environment-provided scalar reward or evaluative label. At each time step the agent receives only a fixed-channel perceptual packet $p_t$ containing channels such as vision, text, audio, proprioception, and sensor values. Quantities such as pain, spice, energy, cognitive error, contact, damage, and actionability are treated as data-driven perceptual dimensions; their valence is inferred from observed transition consequences rather than supplied by the environment.

\OH{} separates prediction, residual dynamics, trajectory evaluation, and policy learning. A neural self-supervised predictor $M_\psi$ learns posterior next-packet expectations $(p_t,a_t)\mapsto \hat p_{t+1}$. The observed packet $p_{t+1}$ yields signed residuals $e_{t+1}=p_{t+1}-\hat p_{t+1}$; a residual-dynamics predictor $D_\omega$ models how those residuals evolve; and a fixed internal trajectory evaluator $\mathcal C_\eta$ maps completed residual trajectories to internal value evidence $Y^{post}$. Its recovery-positive, persistence/growth-negative orientation is grounded in predictive regulation and audited for coefficient-insensitive action ranking. The learner $B_\xi$ is trained from this post-transition evidence and supplies the internal reward/value estimates used for later policy updates and action scoring. The reward-free protocol exposes perceptual transitions while withholding environment scalar rewards, delayed external evaluators, and action-goodness labels.

A conditional error decomposition separates $B_\xi$ evidence-estimation error from residual RL optimization error. In a 2x2-XOR packet task, medicine and chili have opposite learned value under visual XOR contexts; the same pain or spice increase can be positive or negative, while anesthetic lowers pain and cognitive error but has negative learned consequence. $B_\xi$ reaches 0.952 balanced reward-sign accuracy. In a full online-interleaved audit, $M_\psi$ reaches holdout $R^2=0.907$, $B_\xi$ reaches 0.940 sign accuracy, and the policy reaches 0.979 optimal-action accuracy; immediate packet scores, prediction-error rewards, shuffled targets, zero reward, and error-reduction controls collapse under the same audit.
\end{abstract}

\section{Introduction}
Many reinforcement-learning environments expose a scalar reward $r_t$. A first-person agent may instead receive only the next sensory event: pixels change, a sound or text channel changes, a contact sensor changes, energy changes, and motor actionability changes. Such channel values can become value-relevant through consequences while entering the agent only as observations.

We ask whether an agent can learn its own reward-punishment function online from a stream of perceptual events and then use that learned function to acquire a policy. One event is one time step of a fixed-channel sensory bundle,
\begin{equation}
\begin{aligned}
  p_t&=\{(x_t^c,m_t^c)\}_{c\in\mathcal C},\\
  \mathcal C&=\{\text{vision},\text{text},\text{audio},\text{sensor},\text{proprio},\ldots\}.
\end{aligned}
\end{equation}
where $m_t^c$ is a channel mask. The environment transition is
\begin{equation}
  p_{t+1}\sim P_{env}(\cdot\mid p_t,a_t),\qquad \text{reward channel absent}.
\end{equation}
Pain, stress, spice, energy, cognitive error, contact, and actionability are dimensions of $p_t$. Their valence is learned from how actions change posterior expectations and later packet dynamics.

\OH{} implements this setting with role separation. $M_\psi$ is a self-supervised predictor of next-event posterior expectations. Its prediction defines signed residuals $e_{t+1}=p_{t+1}-M_\psi(p_t,a_t)$ once the next packet is observed. $D_\omega$ predicts how residuals evolve and exposes residual-surprise evidence $\zeta$. $\mathcal C_\eta$ evaluates completed post-transition residual trajectories into internal evidence targets $Y^{post}$. $B_\xi$ is trained on those targets after evidence is observed. During future action selection or no-leakage audits, only predictions learned from past transitions may be used; in online RL updates, the same learned estimator can assign an internal reward after the observed transition, exactly as ordinary RL updates use rewards after acting. Raw sensor dimensions, prediction error, curiosity scores, external evaluators, and fixed one-step scalarizations appear as controls.

The evidence emphasizes auditable reward-free controls: the agent receives perceptual transitions while scalar rewards, delayed scalar evaluators, and action-goodness labels remain outside the training interface. We distinguish this from the agent's own internal evaluator: $\mathcal C_\eta$ is constructed after observing consequences and is used to produce learning targets, not to select the action that generated those consequences. In online RL experiments, internal reward assignment is post-transition; in decision-time/no-leakage audits, action scores are predicted from current packet/action variables only. The core nonlinear task uses a 2x2 visual packet in which medicine value depends on $v_{00}\oplus v_{11}$ and chili value depends on $v_{01}\oplus v_{10}$. Identical immediate pain or spice changes receive opposite reward-punishment signs under different visual contexts. We then run an online-interleaved audit where $M_\psi$, distributional $B_\xi$, and the Q policy are updated in the same no-reward stream, closing the loop from interaction to policy learning.

Our contributions are: (1) a fixed-channel event-stream formulation in which the environment updates $p_t$ but emits no environment scalar reward; (2) an online IRL/RL algorithm with an explicit signal-access protocol separating $M_\psi$ posterior prediction, $D_\omega$ residual-dynamics prediction, $\mathcal C_\eta$ post-transition trajectory evaluation, distributional $B_\xi$ internal value-evidence learning, and RL policy learning; (3) a conditional error-decomposition statement that locates policy loss in $B_\xi$ evidence-estimation error and RL optimization error under explicit identifiability, coverage, and realizability conditions; (4) posterior-residual internal targets aligned with the released implementation's no-reward packet stream; (5) non-identifiability audits separating pain, spice, energy, and error-reduction from learned reward labels; and (6) integrated online and family-level controls showing that the chain learns contextual policies without environment rewards or external evaluators.

\section{Method}
\paragraph{Event state and role separation.}
The agent observes fixed-channel packets $p_t$. Channels can include energy, pain, stress, spice, cognitive error, actionability, contact, and damage, but these dimensions are observations rather than reward slots. Their valence is inferred from action-conditioned posterior-residual trajectory evidence.

\begin{figure*}[!t]
\centering
\includegraphics[width=0.88\textwidth]{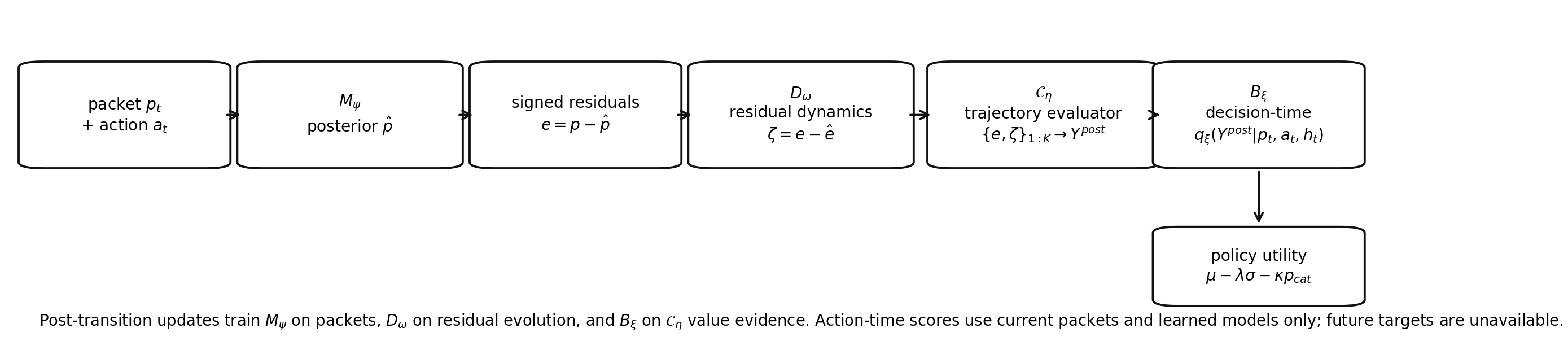}
\caption{Role-separated event-stream learning. The environment returns $p_{t+1}$ and no scalar reward. $M_\psi$ predicts posterior next-packet expectations; $D_\omega$ predicts residual dynamics and exposes residual-surprise evidence $\zeta$; $\mathcal C_\eta$ evaluates completed post-transition residual trajectories into internal value evidence $Y^{post}$. $B_\xi$ is trained from observed evidence windows and provides learned internal reward/value estimates for later policy updates and action scoring. At action time, future $e$, $\zeta$, and $Y^{post}$ are unavailable.}
\label{fig:architecture}
\end{figure*}

\begin{table}[t]
\centering\scriptsize
\setlength{\tabcolsep}{2.5pt}
\begin{tabular}{@{}llll@{}}
\toprule
Part & Learn. & Signal & Timing \\
\midrule
$M_\psi$ & yes & next-packet MSE & post-transition \\
$D_\omega$ & yes & residual-next MSE & post-window \\
$\mathcal C_\eta$ & no & fixed map & stop-gradient \\
$B_\xi$ & yes & fit $Y^{post}$ & update/scoring \\
Policy/Q & yes & internal return & TD/PPO \\
\bottomrule
\end{tabular}
\caption{Component-level training and gradient boundaries. $\mathcal C_\eta$ is a fixed post-transition evaluator. $M_\psi$, $D_\omega$, $B_\xi$, and the policy/value learner are trained by separate online objectives; no gradients pass through the environment, action sampling, or $\mathcal C_\eta$.}
\label{tab:module_training}
\end{table}

The self-supervised predictor learns posterior next-packet expectations,
\begin{equation}
L_M(\psi)=\left\|M_\psi(p_t,a_t)-p_{t+1}\right\|_2^2,
\label{eq:mpsi}
\end{equation}
and defines the signed residual
\begin{equation}
 e_{t+1}=p_{t+1}-\hat p_{t+1},\qquad \hat p_{t+1}=M_\psi(p_t,a_t).
\label{eq:residual}
\end{equation}
A residual's value evidence is determined by direction, persistence, recovery, and downstream controllability; large magnitude by itself marks mismatch or uncertainty. The post-transition evaluator therefore uses a residual-history feature
\begin{equation}
 z_t=\Phi(p_{t:t+K},a_t,\hat p_{t+1:t+K},e_{t+1:t+K},\zeta_{t+2:t+K}),
\label{eq:z_feature}
\end{equation}
where $\zeta$ denotes residual-surprise evidence from a distinct residual-dynamics predictor $D_\omega$, e.g.,
\begin{equation}
\hat e_{t+1}=D_\omega(e_t,p_t,a_t,h_t),\qquad \zeta_{t+1}=e_{t+1}-\hat e_{t+1}.
\end{equation}
After residuals are observed, $D_\omega$ is trained by residual-next MSE, $\sum_{k=1}^{K-1}\|D_\omega(e_{t+k},a_t,k)-e_{t+k+1}\|_2^2$.
$B_\xi$ learns a conditional distribution over internal value evidence. Depending on the audit, $z_t$ may denote the completed post-transition evidence feature used for supervised update, or the pre-decision packet/action/history feature used for held-out action scoring:
\begin{equation}
B_\xi(z_t)=q_\xi(Y_t^{post}\mid z_t),
\label{eq:dist_bxi}
\end{equation}
with heads for mean, lower quantiles, negative-event probability, and positive/negative evidence mass. Point-estimator runs minimize $(B_\xi(z_t)-\operatorname{sg}[Y_t^{post}])^2$, with stop-gradient through the target; distributional runs use the corresponding negative log-likelihood or quantile/heads loss for $q_\xi$. In scalar audits the risk-neutral mean is used; in distributional audits the policy utility is
\begin{equation}
U_\xi(z_t)=\mu_\xi(z_t)-\lambda_\sigma\sigma_\xi(z_t)-\lambda_c P_\xi(Y_t^{post}< -\tau\mid z_t).
\label{eq:risk_utility}
\end{equation}
RL updates from the learned internal utility under the reward-free protocol. Q rows use the TD loss with post-transition $r_t^{int}$ from $B_\xi$ or target evidence; PPO rows use the standard clipped objective with the same internal reward stream. Gradients are not propagated through the environment, action sampling, or $\mathcal C_\eta$; $M_\psi$, $D_\omega$, $B_\xi$, and the policy/value learner have separate online objectives. The timing is explicit: $Y^{post}$ is produced only after the consequence window is observed, while action selection can use only previously learned estimates. Raw prediction loss remains reserved for self-supervised modeling.

\paragraph{Algorithmic contribution.}
\OH{} enforces the signal-access discipline summarized in Table~\ref{tab:module_training}. At action time the agent may use the current packet, policy state, and current parameters of $M_\psi$, $D_\omega$, and $B_\xi$. After the transition, $M_\psi$ receives $p_{t+1}$, $D_\omega$ updates on residual evolution, the residual window becomes observable, and $\mathcal C_\eta$ produces $Y_t^{post}$. $B_\xi$ then learns from this post-transition evidence; online RL can use the assigned internal reward after the transition, while pre-decision audits use only $B_\xi$ scores from current $(p_t,a_t,h_t)$. The controls test prediction loss versus reward, sensor deltas versus valence, shuffled/immediate targets versus posterior-residual value, and coverage versus hard-coded policy.

\paragraph{Posterior-residual internal target.}
The event-stream target is constructed inside the agent from posterior-residual evidence. It is an internal post-transition evaluative target, not an environment-returned reward slot. The orientation of $\mathcal C_\eta$ is a residual-regulation prior: after an action-induced perturbation, residual recovery is positive evidence that the packet stream returns toward a predictable regime, while residual persistence or growth is negative evidence of unresolved mismatch. This direction agrees with predictive-state and homeostatic views of regulation \cite{friston2010free,keramati2011homeostatic} and is audited empirically below. Let
\begin{equation}
\tau_t^{a}=\{p_{t:t+K}^{a},\hat p_{t+1:t+K}^{a},e_{t+1:t+K}^{a}\}
\end{equation}
be the observed or model-predicted packet/residual trajectory after action $a$. The released pure-residual experiments instantiate this orientation as a fixed algebraic map over residual norms. With $n_k=\|e_{t+k}\|_2$,
\begin{equation}
\begin{aligned}
\mathcal C_\eta(\tau)=&\;1.35(n_1-n_K)
-0.38\frac{1}{K-1}\sum_{k=2}^{K} n_k\\
&-0.75[n_K-n_1]_+.
\end{aligned}
\label{eq:ceta_generic}
\end{equation}
\emph{$\mathcal C_\eta$ encodes one prior--recovery positive, persistence/growth negative--and no family rule, success label, goal distance, or action-goodness label.}
The coefficients are fixed across families and actions in the released artifact and are not selected from environment reward, task labels, or oracle actions. A coefficient-origin audit in the supplement varies the numeric recipe while preserving or reversing the orientation. Equal-unit standardized, raw-equal, and random-monotone orientations preserve 0.925, 0.935, and 0.922 of the released top-action rankings, while sign inversion preserves 0.000. The stable information is therefore the monotone direction--recovery positive, persistence/growth negative--rather than exact coefficient magnitudes. The main target is the no-op-normalized posterior-residual advantage
\begin{equation}
Y_t^{post}=\mathcal C_\eta(\tau_t^{do(a_t)})-\mathcal C_\eta(\tau_t^{noop}).
\label{eq:post_adv}
\end{equation}
The no-op branch makes the target an advantage over passive continuation from the same initial event. The $\mathcal C_\eta$ computation occurs after packet observations inside the agent/evaluation code, while action selection uses the learned $B_\xi$ estimate. The artifact includes a static genericity audit of the $\mathcal C_\eta$ implementation: the target function reads residual arrays and algebraic statistics, while family names, action names, visual rules, channel semantics, and oracle labels appear outside target computation.

When intermediate residual evidence is accumulated, Eq.~\ref{eq:post_adv} can be generalized to
\begin{equation}
Y_t^{post,sum}=\sum_{k=0}^{K}\gamma^k\left[\mathcal C_\eta(\tau_{t,k}^{do(a_t)})-\mathcal C_\eta(\tau_{t,k}^{noop})\right].
\label{eq:sum_adv}
\end{equation}
We keep Eq.~\ref{eq:post_adv} as the main experimental target because it matches the released no-op-baseline implementation, and report Eq.~\ref{eq:sum_adv} as a compatible extension.

\begin{algorithm}[!tbp]
\caption{Reward-Free Online Event-Stream IRL/RL}
\begin{algorithmic}[1]
\STATE Observe fixed-channel event $p_t$.
\STATE Choose $a_t$ from the current policy, using only current $p_t$, $M_\psi$, $D_\omega$, $B_\xi$, and policy state.
\STATE Environment returns $p_{t+1}$ and protocol metadata; reward/evaluator/action-label fields remain empty.
\STATE Update $M_\psi$ from $(p_t,a_t)\mapsto p_{t+1}$ by Eq.~\ref{eq:mpsi}.
\STATE Compute signed residual $e_{t+1}=p_{t+1}-M_\psi(p_t,a_t)$; update $D_\omega$ on residual evolution and store $\zeta$ evidence.
\STATE After the evidence window accrues, evaluate trajectories with $\mathcal C_\eta$ and construct $Y_t^{post}$ by Eq.~\ref{eq:post_adv}; it is unavailable before action.
\STATE Update distributional $B_\xi$ toward $Y_t^{post}$ samples or summaries by Eq.~\ref{eq:dist_bxi}.
\STATE Update the RL policy using the learned internal reward/value estimate from $B_\xi$.
\STATE Log success metrics, simulator labels, public rewards, external evaluators, and future targets in evaluation/audit records.
\end{algorithmic}
\label{alg:ohirl}
\end{algorithm}

\paragraph{Signal discipline.}
The implementation records a compact signal-access audit for each experiment. For pre-decision action scoring and held-out no-leakage audits, action scores use current $p_t$, candidate $a_t$, the current $M_\psi$ posterior predictor, the current $D_\omega$ residual-dynamics predictor when present, the current $B_\xi$ utility, and the policy state. For online RL updates, the internal reward assignment is post-transition: after $M_\psi$ trains on $(p_t,a_t)\mapsto p_{t+1}$ and $D_\omega$ trains on residual evolution, $\mathcal C_\eta$ evaluates completed posterior-residual windows and $B_\xi$ trains to predict those value-evidence targets. Success labels, public rewards, simulator labels, and future-window targets are logged as audit/evaluation fields.

\subsection{Conditional Error Decomposition}
The statement records how errors propagate once the internal evidence target is identifiable on the covered support. Its three conditions are explicit. First, \emph{identifiability}: the ideal internal reward-punishment function $B^\star_{post}$ is recoverable from packet histories, posterior residuals, residual-evolution evidence, and post-transition trajectory evaluation. Second, \emph{coverage}: exploration visits the state-action regions on which the learned policy may act. Third, \emph{realizability}: the $B_\xi$ function class can approximate $B^\star_{post}$ on that covered region. Under these conditions, the bound separates internal-evidence estimation error from policy-optimization error.

\paragraph{Proposition.}
Consider a discounted event-stream control problem with discount $\gamma\in[0,1)$. Assume that on the covered region
\begin{equation}
\|B_\xi-B^\star_{post}\|_\infty \le \epsilon_R .
\label{eq:r_error}
\end{equation}
Let $\pi_\theta$ be the policy returned by the RL optimizer using $B_\xi$, and assume its optimization error under $B_\xi$ is bounded by
\begin{equation}
V^{\pi^\star_{\theta}}_{B_\xi}(p)-V^{\pi_\theta}_{B_\xi}(p)
\le \frac{2\epsilon_Q}{1-\gamma},
\label{eq:q_error}
\end{equation}
where $\pi^\star_{\theta}$ is optimal for the learned reward $B_\xi$. Then for the ideal internal posterior-residual reward-punishment function $B^\star_{post}$,
\begin{equation}
V^{\pi^\star}_{B^\star_{post}}(p)-V^{\pi_\theta}_{B^\star_{post}}(p)
\le \frac{2\epsilon_R}{1-\gamma}+\frac{2\epsilon_Q}{1-\gamma},
\label{eq:policy_bound}
\end{equation}
where $\pi^\star$ is optimal for $B^\star_{post}$.

\paragraph{Proof.}
For any fixed policy $\pi$, replacing $B^\star_{post}$ with $B_\xi$ changes the discounted return by at most the discounted sum of per-step reward-punishment estimation errors:
\begin{equation}
\bigl|V^{\pi}_{B_\xi}(p)-V^{\pi}_{B^\star_{post}}(p)\bigr|
\le \sum_{t=0}^{\infty}\gamma^t\epsilon_R
=\frac{\epsilon_R}{1-\gamma}.
\end{equation}
Therefore
\begin{align}
V^{\pi^\star}_{B^\star_{post}}(p)-V^{\pi_\theta}_{B^\star_{post}}(p)
&\le V^{\pi^\star}_{B_\xi}(p)-V^{\pi_\theta}_{B_\xi}(p)+\frac{2\epsilon_R}{1-\gamma} \\
&\le V^{\pi^\star_\theta}_{B_\xi}(p)-V^{\pi_\theta}_{B_\xi}(p)+\frac{2\epsilon_R}{1-\gamma} \\
&\le \frac{2\epsilon_Q}{1-\gamma}+\frac{2\epsilon_R}{1-\gamma}.
\end{align}
The decomposition applies on the support where the internal reward-punishment function is identifiable and covered. The experiments then target the two quantities in the bound: posterior-residual evidence and coverage reduce $\epsilon_R$, while action-before-target ordering and standard RL updates define the policy-optimization term.
\section{Related Work}
\paragraph{Intrinsic rewards.}
Intrinsic motivation methods such as ICM, RND, Plan2Explore, DIAYN, E3B, and URLB encourage exploration through prediction error, novelty, disagreement, information gain, skill diversity, or state coverage \cite{pathak2017icm,burda2019rnd,sekar2020plan2explore,eysenbach2019diayn,henaff2023e3b,laskin2022cic}. Recent systems work also emphasizes that intrinsic-reward results depend strongly on implementation details and standardized evaluation \cite{yuan2024rlexplore}. Classic intrinsic-motivation work already distinguished novelty, prediction progress, and self-generated goals \cite{schmidhuber1991curious,oudeyer2007intrinsic,bellemare2016unifying,tang2017exploration}. A central behavioral trap is that prediction-error curiosity can prefer unpredictable observations instead of useful consequences. Tinker, Doya, and Tani study this issue under a Free-Energy-Principle formulation and show that prediction-error curiosity is vulnerable to observational-noise traps, while hidden-state curiosity is more robust \cite{tinker2024intrinsic}. More recent noise-robust curiosity work likewise rewards learning progress or prediction-error improvement instead of raw error \cite{hou2025lpm,bhaskara2026curiositycritic}. \OH{} uses $M_\psi$ for self-supervised next-packet prediction and empirically separates this modeling loss from the policy reward.

\paragraph{Reward-free and unsupervised RL.}
Reward-free RL studies exploration without task rewards and later planning for reward functions revealed after exploration \cite{jin2020rewardfree,wagenmaker2022rewardfree}. Learning rewards from feedback has also been studied in inverse RL and preference-based RL \cite{ng1999policy,abbeel2004apprenticeship,ziebart2008maxent,christiano2017preferences,du2024porlhf}. Unsupervised RL benchmarks and skill-learning methods study reward-free pretraining followed by downstream task reward evaluation \cite{laskin2022cic,eysenbach2019diayn}. \OH{} differs in protocol: the agent learns an internal reward-punishment estimator $B_\xi$ from $\mathcal C_\eta$ value evidence computed on posterior-residual trajectories and uses it online for policy learning before any downstream task reward is supplied.

\paragraph{World models, active inference, and endogenous signals.}
World-model agents such as DreamerV3 demonstrate the utility of learned dynamics models for control \cite{hafner2023dreamerv3}. Active inference separates epistemic and pragmatic quantities under outcome preferences \cite{friston2010free,friston2017active}. Homeostatic RL studies regulation under internal variables and physiological needs \cite{keramati2011homeostatic,juechems2019where}; this literature motivates the first-person packet formulation, while $B_\xi$ supplies learned valence from $\mathcal C_\eta$ posterior-residual value evidence. In our protocol, first-person sensor changes are data, $M_\psi$ learns posterior packet expectations, and signed value is learned by $B_\xi$ without an environment-authored scalar reward.

\section{Experiments}
The main experiments share a reward-free interface: $env.step(a_t)$ returns observations, packet transitions, termination flags, and diagnostic metadata, while scalar reward, task-success labels, evaluator targets, and oracle action labels are unavailable to model and policy updates. We order the evidence from externally recognizable controls to mechanism diagnostics. The main evidence is: (i) hidden-reward control on public Gymnasium tasks, (ii) a corrected public-data no-leakage action-scoring audit, (iii) a module-role ablation testing whether $\mathcal C_\eta$ can be replaced by prediction error or residual-dynamics proxies, and (iv) posterior-residual mechanism diagnostics on controlled packet families. The packet-family experiments are used to isolate identifiability and failure modes, rather than as the sole evidence base.

\paragraph{Experimental design and fairness.}
All fair baselines operate under the same reward-free interface. They may use observable state or packet transitions, because those are observations returned by the environment, but they do not receive scalar reward, success labels, goal-distance shaping, or action-goodness labels. Consequence-derived baselines such as prediction-error curiosity, learning progress, residual minimization, and residual-of-residual minimization are trained from past transitions and scored from the same decision-time information as OHIRL in no-leakage audits. Reward-provided controllers are reported separately as reference rows because they solve a different interface. The core packet audit uses 50 seeds and 1820 reward-free transitions per seed; logs store action-time packets, posterior-residual targets, model updates, and protocol flags.

\subsection{Standard Hidden-Reward Controls}
We first test whether OHIRL remains useful when the observation stream comes from public environments rather than a procedurally generated packet family. In these controls, the standard environment reward is withheld from same-protocol training. OHIRL observes only environment observations and transitions, constructs internal post-transition evidence, and updates the policy from its learned internal signal. This protocol allows decoded observation-transition facts already present in the observation, but excludes simulator reward, success labels, and hand-coded action values.

\begin{table}[!tbp]
\centering\scriptsize
\setlength{\tabcolsep}{3.0pt}
\resizebox{\columnwidth}{!}{%
\begin{tabular}{lccc}
\toprule
Environment / method & Eval. return & Success / solved & Interface \\
\midrule
CartPole OHIRL & \textbf{243.20$\pm$74.41} & \textbf{0.547$\pm$0.304} & hidden reward \\
CartPole pred.-error & 80.80$\pm$27.97 & -- & hidden reward \\
CartPole RND & 62.00$\pm$20.60 & -- & hidden reward \\
CartPole zero & 9.32$\pm$0.13 & -- & hidden reward \\
\midrule
Taxi OHIRL & \textbf{7.19$\pm$1.08} & \textbf{0.999$\pm$0.006} & hidden reward \\
Taxi transition-error & -1263.29 & -- & hidden reward \\
Taxi RND & -1282.99 & -- & hidden reward \\
Taxi zero & -771.93 & -- & hidden reward \\
\bottomrule
\end{tabular}}
\caption{Hidden-reward public-environment controls over 30 seeds. Rewards are withheld from same-protocol training; returns and success are evaluation metrics.}
\label{tab:hidden_reward_controls}
\end{table}

Table~\ref{tab:hidden_reward_controls} tests the hidden-reward interface: whether an agent can recover policy-useful internal value evidence from observation-transition consequences when the scalar reward channel is withheld. CartPole return is the primary continuous-control metric; solved fraction is a thresholded diagnostic under the same short online hidden-reward budget. The supplement reports the full audit records and protocol flags.

\subsection{Corrected Public-Data No-Leakage Consequence Audit}
We next evaluate decision-time action scoring on public handwritten-digit visual packets. The public dataset defines the observation distribution only; digit labels are not used as reward, evaluator target, or action label. Five non-isomorphic consequence families generate residual trajectories from these contexts, including same-event tail-risk competition, positive/negative component coexistence, delayed masking, low-$\zeta$ stable bad states, and high-$\zeta$ reconstruction. In 25 paired family/seed settings, every fair method learns its score from $(p_t,a_t)$ on training consequences and uses only $(p_t,a_t)$ on held-out images. The strongest distributional $B_\xi$ row, the two-head positive/negative variant, reaches 0.466 risk-optimal action accuracy and 0.452 risk regret, compared with 0.190/0.761 for learned learning-progress, 0.119/0.885 for learned prediction-error curiosity, 0.250/0.772 for learned residual-of-residual minimization, and 0.196/0.824 for shuffled $B_\xi$. Point $B_\xi$ reaches a similar 0.461/0.451. This audit separates OHIRL from residual or novelty shortcuts under equal pre-decision information.

\subsection{Module-Role Ablation}
A dedicated module-role ablation tests whether $\mathcal C_\eta$ can be replaced by $M_\psi$ or $D_\omega$ proxies. Over five residual families and 30 seeds per family, $B_\xi$ trained on $\mathcal C_\eta$ evidence reaches 0.688$\pm$0.145 optimal-action accuracy, compared with 0.413$\pm$0.090 for $M_\psi$ prediction-error targets, 0.253$\pm$0.101 and 0.210$\pm$0.096 for $D_\omega$ predictability and surprise targets, 0.274$\pm$0.142 for shuffled $\mathcal C_\eta$, and 0.207$\pm$0.042 for zero/rest. Thus $D_\omega$ and $\mathcal C_\eta$ are functionally non-substitutable: $D_\omega$ models residual dynamics, while $\mathcal C_\eta$ supplies trajectory-level value evidence for $B_\xi$. A coefficient-origin audit compares the rankings induced by alternative monotone orientations: equal-unit, raw-equal, and random-monotone targets preserve 0.925, 0.935, and 0.922 of the released $\mathcal C_\eta$ top-action rankings, while sign inversion preserves 0.000.

\subsection{Mechanism Diagnostic: XOR Perceptual-Packet Reward-Punishment Learning}
Each packet has a 2x2 binary visual channel $[v_{00},v_{01};v_{10},v_{11}]$ and sensor dimensions including pain, spice, energy, cognitive error, actionability, and tissue damage. Medicine is beneficial when $v_{00}\oplus v_{11}=1$ and harmful when $v_{00}\oplus v_{11}=0$. Chili is beneficial when $v_{01}\oplus v_{10}=1$ and harmful when $v_{01}\oplus v_{10}=0$.

\begin{table*}[!t]
\centering\scriptsize
\resizebox{0.96\textwidth}{!}{%
\begin{tabular}{llcrrrr}
\toprule
Probe & Visual packet & Action & $\Delta$pain & $\Delta$spice & $\Delta$error & Target $Y$ / MLP $B_\xi$ \\
\midrule
Medicine XOR1 therapy & $[[1,0],[0,0]]$ & take medicine & $+0.227$ & $0.000$ & $+0.090$ & $+0.924$ / $+0.865$ \\
Medicine XOR0 poison & $[[0,0],[0,0]]$ & take medicine & $+0.227$ & $0.000$ & $+0.090$ & $-1.626$ / $-1.612$ \\
Chili XOR1 tolerant & $[[0,1],[0,0]]$ & eat chili & $+0.207$ & $+0.550$ & $+0.030$ & $+0.359$ / $+0.345$ \\
Chili XOR0 sensitive & $[[0,0],[0,0]]$ & eat chili & $+0.207$ & $+0.550$ & $+0.030$ & $-1.353$ / $-1.363$ \\
Anesthetic trap & $[[1,1],[0,1]]$ & take anesthetic & $-0.320$ & $0.000$ & $-0.350$ & $-0.923$ / $-0.911$ \\
\bottomrule
\end{tabular}}
\caption{XOR perceptual-packet sign audit. The same immediate pain/spice/error direction can receive opposite delayed reward-punishment signs depending on visual context.}
\label{tab:xor_sign_audit}
\end{table*}

Table~\ref{tab:xor_sign_audit} is the central non-identifiability audit. The medicine rows have identical pain and error changes, yet opposite learned value. The chili rows have identical pain and spice changes, yet opposite learned value. Anesthetic lowers pain and cognitive error and still has negative learned value. Thus pain, spice, and error are dimensions of $p_t$; their reward-punishment sign is learned by $B_\xi$ from visual context and posterior-residual consequences.

\begin{figure*}[!t]
\centering
\includegraphics[width=0.96\textwidth]{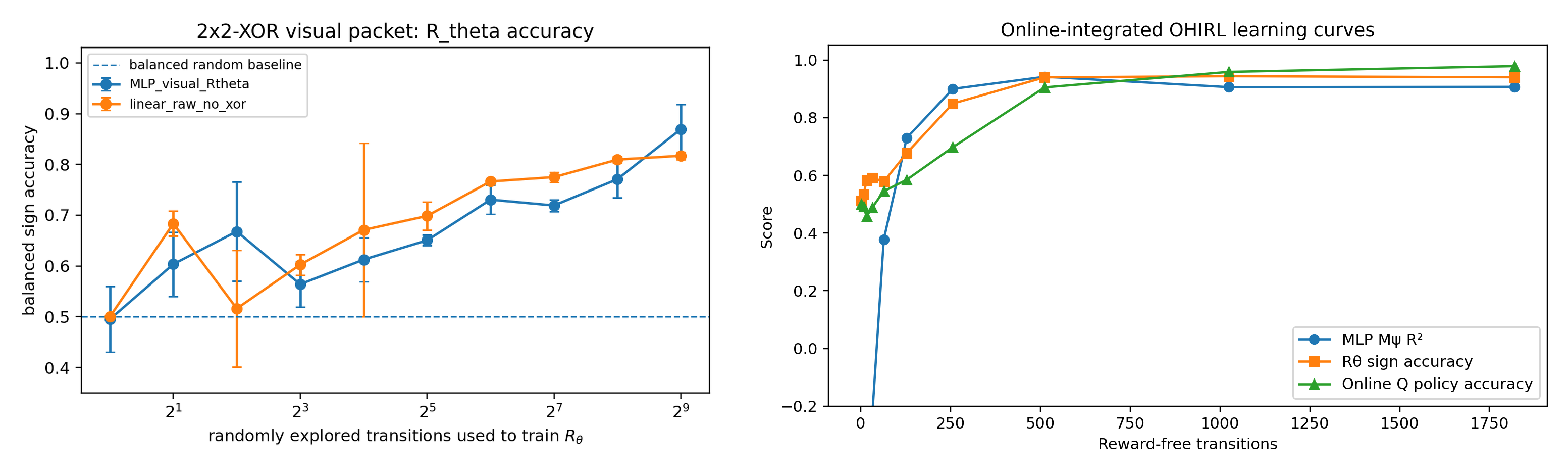}
\caption{Left: MLP $B_\xi$ reward-sign accuracy improves as exploration supplies transitions; accuracy starts near random and reaches 0.952 after 1800 transitions. Right: in the online-interleaved MLP pipeline, $M_\psi$, $B_\xi$, and the Q policy improve in the same reward-free transition stream.}
\label{fig:xor_learning}
\end{figure*}

\subsection{Whole-Pipeline Online Audit}
The sign audit proves that $B_\xi$ can learn context-dependent reward-punishment signs. We next test the complete chain in one stream. For each transition, neural MLP $M_\psi$ is updated self-supervised on $(p_t,a_t)\mapsto p_{t+1}$, neural MLP $B_\xi$ is updated on posterior-residual no-op-advantage targets, and Q-learning updates from the learned $B_\xi$ output. The audit records online interleaving, action-before-target ordering, transition-stream updates, and target-access flags.

\begin{table*}[!t]
\centering\scriptsize
\resizebox{0.96\textwidth}{!}{%
\begin{tabular}{lccc}
\toprule
Agent & Optimal action accuracy & Chosen internal $Y^{post}$ & Anesthetic rate \\
\midrule
Full online OHIRL: MLP $M_\psi$ + $B_\xi$ + online Q & \textbf{0.979$\pm$0.035} & \textbf{0.457$\pm$0.012} & \textbf{0.000$\pm$0.000} \\
Oracle true $Y$ + Q & 1.000$\pm$0.000 & 0.464$\pm$0.000 & 0.000$\pm$0.000 \\
Shuffled $B_\xi$ target + Q & 0.345$\pm$0.207 & -0.169$\pm$0.319 & 0.090$\pm$0.249 \\
Zero reward Q & 0.250$\pm$0.000 & 0.037$\pm$0.000 & 0.000$\pm$0.000 \\
$M_\psi$ prediction error as reward + Q & 0.006$\pm$0.019 & -0.916$\pm$0.075 & 0.093$\pm$0.063 \\
Immediate packet score + Q & 0.000$\pm$0.000 & -0.930$\pm$0.000 & 1.000$\pm$0.000 \\
Cognitive-error reduction + Q & 0.000$\pm$0.000 & -0.930$\pm$0.000 & 1.000$\pm$0.000 \\
\bottomrule
\end{tabular}}
\caption{Whole-pipeline online audit over 50 seeds. Full OHIRL updates MLP $M_\psi$, MLP $B_\xi$, and online Q in the same reward-free transition stream.}
\label{tab:online_integrated}
\end{table*}

At the final checkpoint, MLP $M_\psi$ reaches next-packet holdout $R^2=0.907\pm0.006$, MLP $B_\xi$ reaches balanced reward-sign accuracy $0.940\pm0.037$, and the online Q policy reaches $0.979\pm0.035$ optimal-action accuracy. Table~\ref{tab:online_integrated} shows prediction-error reward at 0.006 accuracy, shuffled $B_\xi$ targets at 0.345, and cognitive-error reduction selecting anesthetic in every seed. Full OHIRL's advantage comes from the learned reward-punishment estimator rather than prediction loss or local sensor improvement.

\subsection{Environment-Family Generalization}
We next test five reward-free perceptual-packet families that change the visual rule determining whether medicine or chili has positive posterior-residual consequence: diagonal XOR, row XOR, column XOR, mixed parity, and noisy-switch XOR. Each family keeps the same reward-free interface and uses online-interleaved MLP $M_\psi$ + MLP $B_\xi$ + Q learning. The protocol begins with a stratified coverage phase and then switches to epsilon-greedy online control. Each family is evaluated over 50 seeds with 800 reward-free transitions per seed.

\begin{table*}[!t]
\centering\scriptsize
\setlength{\tabcolsep}{2.7pt}
\resizebox{0.98\textwidth}{!}{%
\begin{tabular}{lcccccc}
\toprule
Family & OHIRL acc. & Oracle acc. & Regret & Shuffled $B_\xi$ & Zero reward & $M_\psi$-err. / immediate / err.-red. \\
\midrule
col XOR & 0.893 & 1.000 & 0.035 & 0.273 & 0.250 & 0.006 / 0.000 / 0.000 \\
diag XOR & 0.916 & 1.000 & 0.028 & 0.289 & 0.250 & 0.005 / 0.000 / 0.000 \\
noisy-switch XOR & 0.879 & 1.000 & 0.039 & 0.316 & 0.188 & 0.006 / 0.000 / 0.000 \\
parity-mixed & 0.908 & 1.000 & 0.031 & 0.299 & 0.313 & 0.011 / 0.000 / 0.000 \\
row XOR & 0.879 & 1.000 & 0.041 & 0.293 & 0.250 & 0.005 / 0.000 / 0.000 \\
\midrule
Aggregate & \textbf{0.895$\pm$0.095} & 1.000 & 0.035 & 0.294 & 0.250 & 0.007 / 0.000 / 0.000 \\
\bottomrule
\end{tabular}}
\caption{Environment-family generalization. Full online OHIRL is positive in every family and remains above prediction-error, immediate-score, zero-reward, shuffled-$B_\xi$, and error-reduction controls.}
\label{tab:family_generalization}
\end{table*}

The learning curves pass a preflight audit: across families, $M_\psi$ $R^2$ starts at $-2.462$ and reaches $0.907$, $B_\xi$ balanced sign accuracy starts at $0.495$ and reaches $0.925$, and policy accuracy starts at $0.500$ and reaches $0.895$. The online-order audit confirms that actions are selected before terminal targets are constructed; the environment returns no reward slot; and optimal-action accuracy, regret, and chosen internal $Y^{post}$ are evaluation metrics only, not training targets.

\subsection{Pure Posterior-Residual Learning Audit}
The preceding family suite uses auditable no-op-normalized posterior-residual advantages saved in the implementation. To test the stronger claim that $B_\xi$ can learn reward-punishment structure directly from $M_\psi$ posterior-error dynamics, we run a separate pure posterior-residual full50 audit. In this audit the environment returns packet transitions and protocol metadata with reward/evaluator/action-label fields unavailable to training. The scalar target for $B_\xi$ is computed by $\mathcal C_\eta$ from the signed residual sequence produced by $M_\psi$ predictions and observed packet transitions using Eq.~\ref{eq:ceta_generic}. Thus the target is a generic internal residual-statistics functional over residual recovery, persistence, and growth; family identity, XOR rules, action labels, and channel-specific reward signs live in the environment dynamics rather than in the target map. The target is still an agent-side post-transition evaluator, so the claim is absence of environment reward, not absence of internal evaluation. The policy updates from the learned $B_\xi$ prediction. The audit uses five nonlinear families, 50 seeds per family, and 512 reward-free transitions per seed under online-interleaved updates with initial stratified coverage.

\begin{table}[!tbp]
\centering\scriptsize
\setlength{\tabcolsep}{3.0pt}
\resizebox{\columnwidth}{!}{%
\begin{tabular}{lccc}
\toprule
Agent & Policy acc. & Regret & Anesthetic \\
\midrule
Full posterior-residual OHIRL & \textbf{0.897$\pm$0.080} & 0.0069 & 0.000 \\
Privileged residual oracle & 0.937 & 0.0017 & 0.000 \\
Prediction-error reward & 0.380 & 0.231 & 0.000 \\
Immediate error minimization & 0.098 & 0.736 & 0.551 \\
Shuffled residual target & 0.221 & 0.466 & 0.254 \\
Zero reward & 0.235 & 0.316 & 0.000 \\
\bottomrule
\end{tabular}}
\caption{Pure posterior-residual full50 audit. $B_\xi$ is trained from $M_\psi$ signed residual trajectories through the released fixed-orientation $\mathcal C_\eta$ in Eq.~\ref{eq:ceta_generic}; the target map reads residual arrays and shared residual statistics.}
\label{tab:pure_residual}
\end{table}

The curve sanity check is also normal: across families, $M_\psi$ $R^2$ starts at $-2.139$ and reaches $0.888$, $B_\xi$ sign accuracy starts at $0.617$ and reaches $0.940$, and policy accuracy starts at $0.247$ and reaches $0.897$. This result supports the posterior-residual mechanism: $M_\psi$ supplies the evidence used by $B_\xi$, and the reward signal comes from learned residual structure.

\paragraph{Mechanism summary.}
The controlled packet-family diagnostics below isolate the posterior-residual mechanism after the public hidden-reward and public-context controls. They are not the sole evidence base; their role is to show why local observation scores, prediction error, cognitive-error reduction, shuffled value evidence, and residual-only proxies fail even when the consequence structure is fully auditable.

\section{Discussion}
The environment updates perceptual dimensions in $p_t$ under a reward-free interface. Pain, pressure, spice, energy, damage, cognitive error, contact, and actionability enter as observations. $M_\psi$ converts these streams into posterior expectations and signed residuals; $D_\omega$ characterizes residual evolution; $\mathcal C_\eta$ assigns internal value evidence to completed post-transition trajectories; and $B_\xi$ learns from those targets to provide internal rewards after observed transitions and predicted value estimates for later action scoring. Replacing $B_\xi$ with raw $M_\psi$ prediction error gives 0.006 optimal-action accuracy in the online audit, separating prediction loss from learned reward.

The error-decomposition result scopes generalization through identifiability, coverage, and realizability: Eq.~\ref{eq:policy_bound} relates $B_\xi$ estimation error and RL optimization error to policy loss under the ideal internal posterior-residual reward-punishment function. Empirically, OHIRL approaches oracle across nonlinear reward-free packet families with online-interleaved learning and an initial stratified coverage phase. The controls separate observation access from reward provision: every method observes the same packet dimensions, while hand-coded packet scores, curiosity, error reduction, shuffled $B_\xi$, and zero reward trail learned posterior-residual value.

\section{Conclusion}
We presented an online IRL/RL formulation in which the environment updates perceptual packets $p_t$ while withholding scalar reward. $M_\psi$ learns posterior next-packet expectations; $D_\omega$ models residual evolution; $\mathcal C_\eta$ evaluates completed post-transition residual trajectories into internal value evidence; $B_\xi$ learns this reward-punishment evidence model; and RL uses learned $B_\xi$ estimates as internal rewards or action scores according to the audited timing protocol. The hidden-reward CartPole/Taxi controls, public-context no-leakage suite, module-role ablation, and posterior-residual mechanism diagnostics show OHIRL learning contextual policies under this event-stream interface, with direct observation scores and reward-like shortcuts trailing or collapsing. The module-role audit supports the functional separation of $D_\omega$ dynamics prediction from $\mathcal C_\eta$ value-evidence evaluation, while distributional heads represent competing or coexisting positive and negative outcomes. The conditional decomposition identifies when posterior-residual value learning controls policy suboptimality through $B_\xi$ estimation error and RL optimization error.

\section{Reproducibility}
The accompanying technical supplement and reproducibility artifact contain scripts, CSV/JSON outputs, reports, and audits for the hidden-reward controls, public-context no-leakage suite, module-role and coefficient-origin audits, packet-family diagnostics, reward-free PPO, public-digit counterexample, and corrected distributional $B_\xi$ experiments. \texttt{RESULT\_SOURCE\_CHECK.md} maps reported results to source files.


\begingroup\footnotesize
\begin{thebibliography}{99}
\bibitem[Burda et~al.(2019)]{burda2019rnd} Burda, Y.; Edwards, H.; Storkey, A.; and Klimov, O. 2019. Exploration by Random Network Distillation. In \emph{ICLR}.
\bibitem[Brockman et~al.(2016)]{brockman2016gym} Brockman, G.; Cheung, V.; Pettersson, L.; Schneider, J.; Schulman, J.; Tang, J.; and Zaremba, W. 2016. OpenAI Gym. \emph{arXiv:1606.01540}.
\bibitem[Chevalier-Boisvert et~al.(2023)]{chevalier2023minigrid} Chevalier-Boisvert, M.; Dai, B.; Towers, M.; Perez-Vicente, R.; Willems, L.; Lahlou, S.; Pal, S.; Castro, P. S.; and Terry, J. 2023. MiniGrid and MiniWorld: Modular Reinforcement Learning Environments. \emph{CoRR}.
\bibitem[Eysenbach et~al.(2019)]{eysenbach2019diayn} Eysenbach, B.; Gupta, A.; Ibarz, J.; and Levine, S. 2019. Diversity Is All You Need: Learning Skills without a Reward Function. In \emph{ICLR}.
\bibitem[Friston(2010)]{friston2010free} Friston, K. 2010. The Free-Energy Principle: A Unified Brain Theory? \emph{Nature Reviews Neuroscience}, 11(2): 127--138.
\bibitem[Friston et~al.(2017)]{friston2017active} Friston, K.; FitzGerald, T.; Rigoli, F.; Schwartenbeck, P.; and Pezzulo, G. 2017. Active Inference: A Process Theory. \emph{Neural Computation}, 29(1): 1--49.
\bibitem[Hafner et~al.(2023)]{hafner2023dreamerv3} Hafner, D.; Pasukonis, J.; Ba, J.; and Lillicrap, T. 2023. Mastering Diverse Domains through World Models. \emph{arXiv:2301.04104}.
\bibitem[Haarnoja et~al.(2018)]{haarnoja2018sac} Haarnoja, T.; Zhou, A.; Abbeel, P.; and Levine, S. 2018. Soft Actor-Critic: Off-Policy Maximum Entropy Deep Reinforcement Learning with a Stochastic Actor. In \emph{ICML}.
\bibitem[Henaff et~al.(2022)]{henaff2023e3b} Henaff, M.; Raileanu, R.; Jiang, M.; and Rocktaschel, T. 2022. Exploration via Elliptical Episodic Bonuses. In \emph{NeurIPS}.
\bibitem[Juechems and Summerfield(2019)]{juechems2019where} Juechems, K.; and Summerfield, C. 2019. Where Does Value Come From? \emph{Trends in Cognitive Sciences}, 23(10): 836--850.
\bibitem[Keramati and Gutkin(2011)]{keramati2011homeostatic} Keramati, M.; and Gutkin, B. 2011. A Reinforcement Learning Theory for Homeostatic Regulation. \emph{Psychological Review}, 118(4): 604--644.
\bibitem[Laskin et~al.(2021)]{laskin2022cic} Laskin, M.; Yarats, D.; Liu, H.; Lee, K.; Zhan, A.; Lu, K.; Cang, C.; Pinto, L.; and Abbeel, P. 2021. URLB: Unsupervised Reinforcement Learning Benchmark. In \emph{NeurIPS Datasets and Benchmarks}.
\bibitem[Pathak et~al.(2017)]{pathak2017icm} Pathak, D.; Agrawal, P.; Efros, A. A.; and Darrell, T. 2017. Curiosity-Driven Exploration by Self-Supervised Prediction. In \emph{ICML Workshop}.
\bibitem[Schulman et~al.(2017)]{schulman2017ppo} Schulman, J.; Wolski, F.; Dhariwal, P.; Radford, A.; and Klimov, O. 2017. Proximal Policy Optimization Algorithms. \emph{arXiv:1707.06347}.
\bibitem[Sekar et~al.(2020)]{sekar2020plan2explore} Sekar, R.; Rybkin, O.; Daniilidis, K.; Abbeel, P.; Hafner, D.; and Pathak, D. 2020. Planning to Explore via Self-Supervised World Models. In \emph{ICML}.
\bibitem[Todorov et~al.(2012)]{todorov2012mujoco} Todorov, E.; Erez, T.; and Tassa, Y. 2012. MuJoCo: A Physics Engine for Model-Based Control. In \emph{IROS}.

\bibitem[Abbeel and Ng(2004)]{abbeel2004apprenticeship} Abbeel, P.; and Ng, A. Y. 2004. Apprenticeship Learning via Inverse Reinforcement Learning. In \emph{ICML}.
\bibitem[Bellemare et~al.(2016)]{bellemare2016unifying} Bellemare, M. G.; Srinivasan, S.; Ostrovski, G.; Schaul, T.; Saxton, D.; and Munos, R. 2016. Unifying Count-Based Exploration and Intrinsic Motivation. In \emph{NeurIPS}.
\bibitem[Bhaskara and Wang(2026)]{bhaskara2026curiositycritic} Bhaskara, V.; and Wang, H. 2026. Curiosity-Critic: Cumulative Prediction Error Improvement as a Tractable Intrinsic Reward for World Model Training. \emph{arXiv:2604.18701}.
\bibitem[Christiano et~al.(2017)]{christiano2017preferences} Christiano, P. F.; Leike, J.; Brown, T.; Martic, M.; Legg, S.; and Amodei, D. 2017. Deep Reinforcement Learning from Human Preferences. In \emph{NeurIPS}.
\bibitem[Du et~al.(2024)]{du2024porlhf} Du, Y.; Winnicki, A.; Dalal, G.; Mannor, S.; and Srikant, R. 2024. Exploration-Driven Policy Optimization in RLHF: Theoretical Insights on Efficient Data Utilization. \emph{arXiv:2402.10342}.
\bibitem[Hou, An, and Du(2025)]{hou2025lpm} Hou, Z.; An, Z.; and Du, W. 2025. Beyond Noisy-TVs: Noise-Robust Exploration via Learning Progress Monitoring. \emph{arXiv:2509.25438}.
\bibitem[Kearns and Singh(2002)]{kearns2002near} Kearns, M.; and Singh, S. 2002. Near-Optimal Reinforcement Learning in Polynomial Time. \emph{Machine Learning}, 49: 209--232.
\bibitem[Jin et~al.(2020)]{jin2020rewardfree} Jin, C.; Krishnamurthy, A.; Simchowitz, M.; and Yu, T. 2020. Reward-Free Exploration for Reinforcement Learning. In \emph{ICML}.
\bibitem[Ng, Harada, and Russell(1999)]{ng1999policy} Ng, A. Y.; Harada, D.; and Russell, S. 1999. Policy Invariance under Reward Transformations: Theory and Application to Reward Shaping. In \emph{ICML}.
\bibitem[Oudeyer and Kaplan(2007)]{oudeyer2007intrinsic} Oudeyer, P.-Y.; and Kaplan, F. 2007. What Is Intrinsic Motivation? A Typology of Computational Approaches. \emph{Frontiers in Neurorobotics}, 1: 6.
\bibitem[Pedregosa et~al.(2011)]{pedregosa2011scikit} Pedregosa, F.; Varoquaux, G.; Gramfort, A.; Michel, V.; Thirion, B.; Grisel, O.; Blondel, M.; Prettenhofer, P.; Weiss, R.; Dubourg, V.; Vanderplas, J.; Passos, A.; Cournapeau, D.; Brucher, M.; Perrot, M.; and Duchesnay, E. 2011. Scikit-learn: Machine Learning in Python. \emph{Journal of Machine Learning Research}, 12: 2825--2830.
\bibitem[Puterman(1994)]{puterman1994mdp} Puterman, M. L. 1994. \emph{Markov Decision Processes: Discrete Stochastic Dynamic Programming}. Wiley.
\bibitem[Schmidhuber(1991)]{schmidhuber1991curious} Schmidhuber, J. 1991. A Possibility for Implementing Curiosity and Boredom in Model-Building Neural Controllers. In \emph{Proc. SAB}.
\bibitem[Sutton and Barto(2018)]{sutton2018rl} Sutton, R. S.; and Barto, A. G. 2018. \emph{Reinforcement Learning: An Introduction}. MIT Press, 2nd edition.
\bibitem[Tang et~al.(2017)]{tang2017exploration} Tang, H.; Houthooft, R.; Foote, D.; Stooke, A.; Chen, X.; Duan, Y.; Schulman, J.; De Turck, F.; and Abbeel, P. 2017. Exploration: A Study of Count-Based Exploration for Deep Reinforcement Learning. In \emph{NeurIPS}.
\bibitem[Tinker, Doya, and Tani(2024)]{tinker2024intrinsic} Tinker, T. J.; Doya, K.; and Tani, J. 2024. Intrinsic Rewards for Exploration without Harm from Observational Noise: A Simulation Study Based on the Free Energy Principle. \emph{arXiv:2405.07473}.
\bibitem[Wagenmaker et~al.(2022)]{wagenmaker2022rewardfree} Wagenmaker, A.; Chen, Y.; Simchowitz, M.; Du, S. S.; and Jamieson, K. 2022. Reward-Free RL is No Harder Than Reward-Aware RL in Linear Markov Decision Processes. In \emph{COLT}.
\bibitem[Yuan et~al.(2024)]{yuan2024rlexplore} Yuan, M.; Castanyer, R. C.; Li, B.; Jin, X.; Berseth, G.; and Zeng, W. 2024. RLeXplore: Accelerating Research in Intrinsically-Motivated Reinforcement Learning. \emph{arXiv:2405.19548}.
\bibitem[Ziebart et~al.(2008)]{ziebart2008maxent} Ziebart, B. D.; Maas, A.; Bagnell, J. A.; and Dey, A. K. 2008. Maximum Entropy Inverse Reinforcement Learning. In \emph{AAAI}.
\end{thebibliography}
\endgroup
\end{document}